\documentclass{vgtc}                          

\ifpdf
  \pdfoutput=1\relax                   
  \usepackage{graphicx}                
  \DeclareGraphicsExtensions{.pdf,.png,.jpg,.jpeg} 
\else
  \usepackage{graphicx}                
  \DeclareGraphicsExtensions{.eps}     
\fi%

\graphicspath{{figures/}{pictures/}{images/}{./}} 

\usepackage{microtype}                 
\PassOptionsToPackage{warn}{textcomp}  
\usepackage{textcomp}                  
\usepackage{mathptmx}                  
\usepackage{setspace}
\usepackage{times}                     

\usepackage{cite}                      
\usepackage{tabu}                      
\usepackage{booktabs}                  
\usepackage{amsmath}
\usepackage{amsfonts}
\usepackage{multirow}
\usepackage{tikz}
\usepackage{tabu}
\usepackage{pgfplots}
\usepackage{subfig}
\usepackage{authblk}
\usepackage[export]{adjustbox}
\usepgfplotslibrary{groupplots}
\pgfplotsset{every tick label/.append style={font=\normalsize},label style=
 {font=\normalsize},width=6.5cm,height=5.5cm,compat=1.3}

\onlineid{1061}

\vgtcinsertpkg

\title{LE-HGR: A Lightweight and Efficient RGB-based Online Gesture Recognition Network for Embedded AR Devices }

\newcommand*\samethanks[1][\value{footnote}]{\footnotemark[#1]}
\author[1]{Hongwei Xie \thanks{These authors contributed equally to this work.}  \qquad Jiafang Wang \samethanks \qquad Baitao Shao \samethanks  \qquad Jian Gu \qquad Mingyang Li\\
{A.I. Labs, Alibaba Group Inc., Hangzhou, China\\
  \tt\small \{hongwei.xhw\textsuperscript{*}, jiafang.wjf\textsuperscript{*}, baitao.sbt\textsuperscript{*}, gujian.gj, mingyangli\}@alibaba-inc.com}
}

\abstract{Online hand gesture recognition (HGR) techniques 
are essential in
augmented reality (AR) applications for enabling natural human-to-computer interaction and communication.
In recent years, the consumer market for low-cost AR devices has been rapidly growing, while 
the technology maturity in this domain is still limited. 
Those devices are typical of low prices, limited memory, and resource-constrained computational units, 
which makes online HGR a challenging problem.
To tackle this problem, 
we propose a lightweight and computationally efficient HGR framework, namely LE-HGR, to enable real-time gesture recognition
on embedded devices with low computing power. We also show that the proposed method is of high accuracy and robustness, which is able to reach high-end performance in a variety of complicated interaction environments.

To achieve our goal, we first propose a cascaded multi-task convolutional neural network (CNN) to 
simultaneously predict probabilities of hand detection and 
regress hand keypoint locations online.
We show that, with the proposed cascaded architecture design, false-positive estimates can be largely eliminated.
Additionally, an associated mapping approach is introduced to track the hand trace via the predicted locations, 
which addresses the interference of multi-handedness. 
Subsequently, we propose a trace sequence neural network (TraceSeqNN) to recognize the hand gesture by exploiting the motion features of the tracked trace. 
Finally, we provide a variety of
experimental results 
to show that the proposed framework is able to achieve state-of-the-art accuracy with 
significantly reduced computational cost, 
which are the key properties for enabling real-time applications in low-cost commercial devices such as mobile devices and AR/VR headsets.

} 

\teaser{
  \centering
  \includegraphics[scale=0.4]{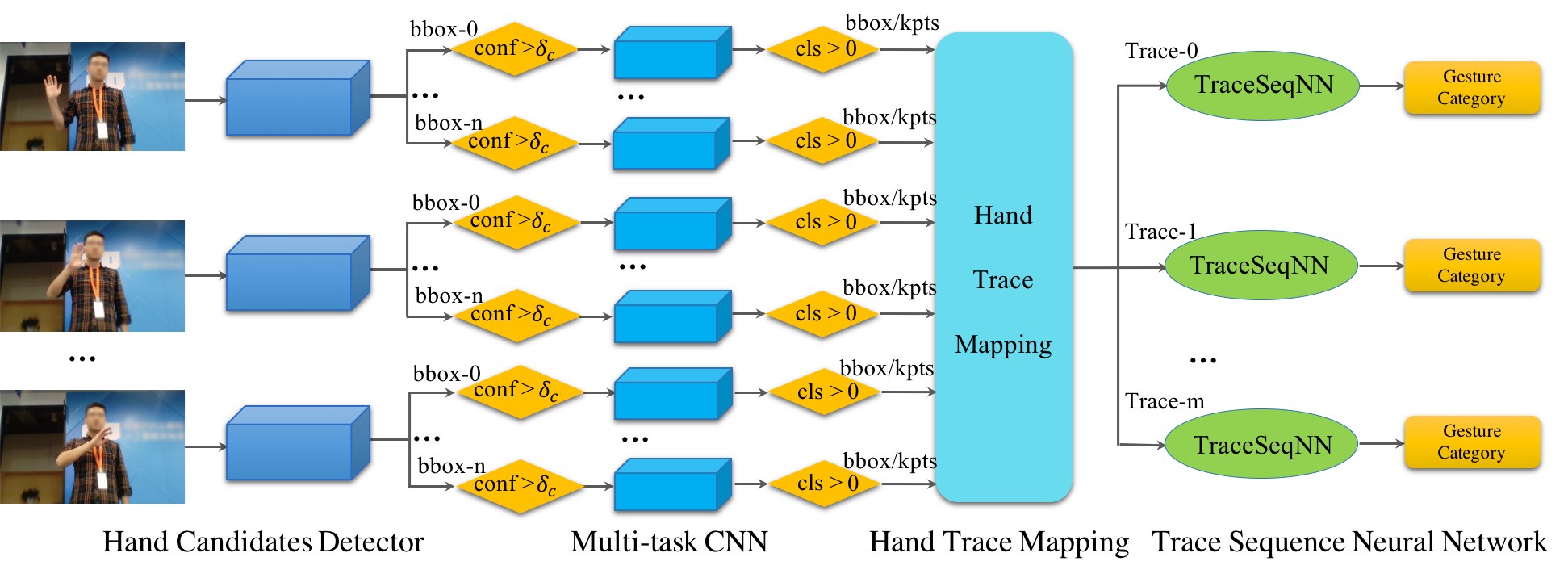}
  \caption{\textbf{LE-HGR Framework.} The overall framework of the proposed gesture recognition algorithm. The framework contains four sub-modules: Hand candidate detector, 
  multi-task CNN, hand trace mapping, and trace sequence network.}
  \label{fig:framework}
}

\begin{document}


\firstsection{Introduction}


\maketitle
In recent years, AR devices and applications have received unprecedented attentions from both 
research community and industry. 
To enable high-quality AR experience, an AR device should be able to 
understand its own motion~\cite{mur2017orb, li2013high, zheng2017photometric}, 
understand its surrounding environments~\cite{klingensmith2015chisel, lin2017feature}, 
and interact with human via control commands~\cite{sagawa2015poster,tewari2017poster}.
Among the interaction technologies, HGR is one of the most natural and 
efficient methods, which greatly enhances the AR experience. Thus,
in this paper, we focus on the problem of HGR, to enable high-quality experience on 
low-cost AR devices.

Early studies on HGR mainly focus on the 3D skeleton-based method by using RGB-D cameras.
RGB-D camera is able to directly measure depths, which greatly simplifies the 
algorithm design. However, its relatively high price makes it not affordable for 
low-cost AR devices.
Recently, Molchanov \textit{et al.} \cite{molchanov2015hand} introduced a RGB-based HGR system
by using a 3D-CNN based method.
While this method is theoretically sound and interesting, the incurred high computational cost makes it infeasible
on low-cost devices. Moreover, the 3D-CNN-based algorithms will suffer from reduced performance in scenes with complex background or with multiple hands.



Although real-time HGR on low-cost embedded devices is a less-explored topic, the applications using embedded-enabled CNNs are under active study by building lightweight efficient networks~\cite{redmon2016you,liu2016ssd,howard2017mobilenets}. However, the well-established techniques for other visual tasks are not directly applicable for gesture recognition problem due to its challenges in high degrees of freedom (DoF) and fast speed of hand moving. This makes real-time hand interaction on low-cost embedded device a challenging problem. 


In this paper, we propose a lightweight and computationally efficient gesture recognition network, namely LE-HGR, to enable real-time gesture recognition on 
embedded devices with limited computational and memory resources, as well as efficiently enhance the generalization ability of 
the complex AR interaction scenes. 
In the first step, we fed low-resolution images into candidate detector for real-time inference. To speed up the inference phase, we adopt the MobileNet algorithm \cite{howard2017mobilenets,sandler2018mobilenetv2} 
as our backbone network to largely compress the required parameters and computations. 
Subsequently, a multi-task CNN is exploited to extract the temporal features of motion trace and the spatial features of hand shape.
Finally, we implement a TraceSeqNN to recognize the hand gesture based on these motion features. 
A video file is also attached as our supplementary material in which real-time gesture recognition is used for AR applications.

\begin{figure}
  \centering
  \includegraphics[width=1.0\linewidth]{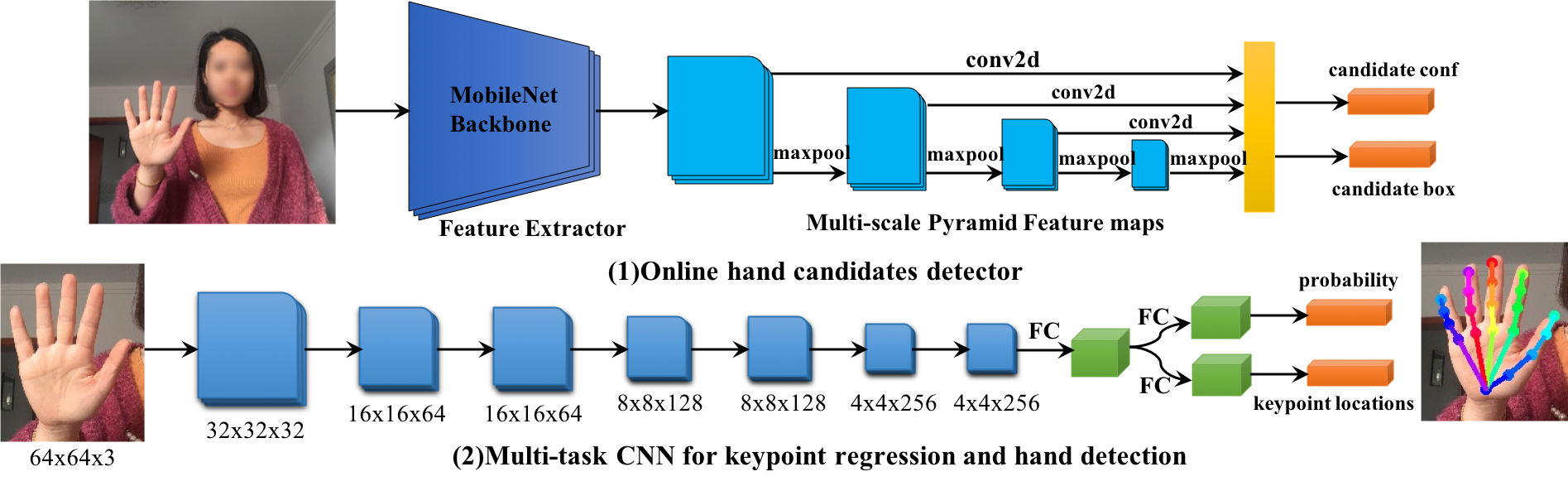}
  \caption{\textbf{Multi-task cascaded network architecture.} Top sub-figure: The online detector for generating
  candidate bounding boxes for hand. Bottom sub-figure: The multi-task CNN for eliminating the false candidates, and regressing the keypoint locations.
  }
  \label{fig:multitask_cascaded_archi}
\end{figure}

Our main contributions are summarized as follows:
\begin{itemize}

  \item Firstly, we design a cascaded multi-task network to enable real-time hand detection on AR devices. Subsequently, to enables interactive hand applications in complicated environments with multiple persons or multiple hands, we propose a fast mapping method to track the hand trace via predicted locations,.
  \item Secondly, to tackle the online HGR, we design the TraceSeqNN to encode the temporal correlation information and predict the gesture category.
  The velocity branch and shape branch are exploited to better learn the motion information. 
  Additionally, an effective data augmentation method for the sequence samples is proposed to prevent overfitting
  and simulate different temporal domains of motion movements.
  \item Finally, we propose a novel gesture recognition framework and apply our system to AR interacting applications for enhancing the AR immersion experiences. Specifically, we show that our method outperforms R3DCNN\cite{gupta2016online} by $1.2\%$ accuracy improvement on Nvidia gesture dataset \cite{gupta2016online}. More importantly, our method runs {\em 500x} faster compared to \cite{gupta2016online}, and is able to perform real-time tasks on low-cost devices. 
\end{itemize}

\begin{figure}
  \centering
  \includegraphics[width=1.0\linewidth]{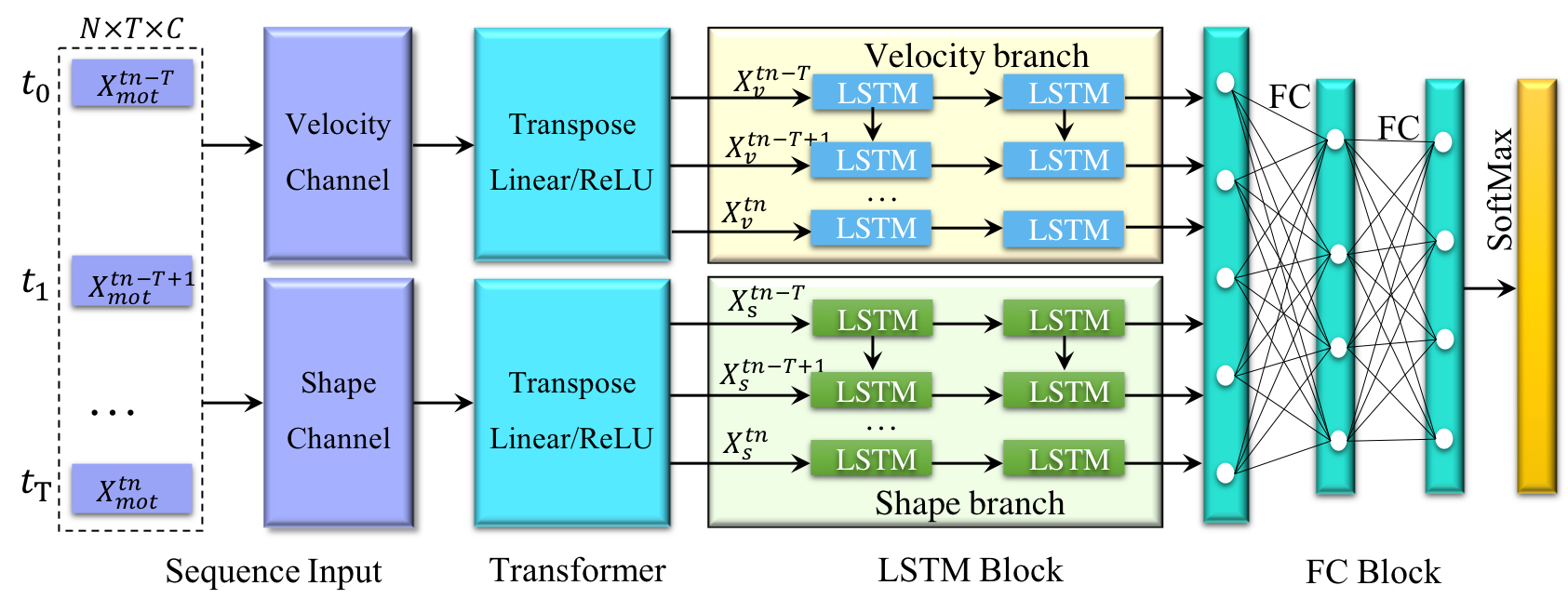}
  \caption{\textbf{The proposed TraceSeqNN architecture.} The TraceSeqNN is composed of transformer, LSTM and FC blocks. We adopt velocity branch and shape branch to better learn the motion features.}
  \label{fig:TraceSeqNN}
\end{figure}

\section{Related Works}

\subsection{Visual Recognition Methods for Hand Detection}

One of the most well known visual recognition algorithm series is the 
R-CNN families\cite{girshick2015fast,ren2015faster}, which generate potential bounding boxes via region proposal methods and run a classifier on these proposed boxes.
However, these methods require a large amount of computational resources, making themselves infeasible for low-cost embedded devices.

Unlike region proposal-based techniques, Single shot multi-box detector (SSD) \cite{DBLP:journals/corr/LiuAESR15} 
produced predictions on multi-scale feature maps and achieves competitive accuracy 
even with the relatively low-resolution input.
To enable convolutional neural network in real-time on low-cost devices, Howard \textit{et.al} \cite{howard2017mobilenets}
proposed MobileNet network, which is based on depth-wise separable convolutions to compress 
the number of parameters. 

\subsection{Gesture Recognition using Spatial-temporal Features}
\subsubsection{HGR via the RGB-D based Approachs}
RGB-D cameras have the natural advantage of action recognition due to its capability of directly capturing 3D dense point cloud data.
In recent years, a variety of methods have been successfully applied to the 3D skeleton-based action recognition problems.
\cite{zhang2019view,zhao2018skeleton,liu2018skeleton,xie2018memory,ohn2014hand,hou2018spatial}.
Smedt \textit{et al.} \cite{de20173d,zhao2018skeleton} encoded the temporal pyramid information using the 
3D skeleton-based geometric features and performed a linear support vector machine (SVM) to achieve a classification task.
Chen \textit{et al.} \cite{chen2017motion} proposed a motion feature augmented recurrent neural network (RNN) for skeleton-based HGR.


\subsubsection{Gesture Recognition using the Sequential Video Sequences}
A large number of 3D-CNN-based methods \cite{wang2018human,miao2017multimodal,tran2015learning,gupta2016online} have been developed 
for producing better performance on various video analysis tasks.
Them, ConvLSTM \cite{xingjian2015convolutional,zhang2017learning,zhu2017multimodal,wang2017large}
combined the 3DCNN and LSTM to effectively learn different levels of spatial-temporal characteristics. CLDNN \cite{sainath2015convolutional} analyzed the effect of adding CNN layers to LSTM and LRCN \cite{donahue2015long} extended the convolutional LSTM for visual recognition problems.
Besides, multiple-stream based methods have been proposed for tackling the multiple modality data \cite{nishida2015multimodal,simonyan2015two},
which separately learn the spatial features from video frames and the temporal feature from the dense optical flow for action recognition.

\section{Methodology}
\subsection{Framework Overview}
In this section, we illustrate the overall pipeline of our approach.
As shown in \autoref{fig:framework}, our LE-HGR mainly consists of four stages: detecting hand candidates from input images, refining detection results and regressing hand keypoints using a multi-task network, trace mapping for addressing multi-hand ambiguities, and gesture recognition from the hand-skeleton sequence. 
At the first stage, the hand bounding boxes are detected via 
an ultra-simplified SSD \cite{liu2016ssd} network. Since this lightweight detector is inevitable to produce false-positive detection results, we propose a cascaded multi-task network to refine those results, as well as regress the hand keypoints in parallel. 
To effectively handle the situation when multiple hands appearing simultaneously in input images, we maintain
the temporal traces via associating the predicted hands to the existed hand motion trails.
Finally, hand keypoints in the temporal trace are stacked as the sequential input 
of the TraceSeqNN to recognize the gesture.

\subsection{Joint Hand Detection and Keypoints Regression using the Multi-task Cascaded Network}
In this section, we describe the architecture and training objective of our multi-task network,
which is also shown in \autoref{fig:multitask_cascaded_archi}.

\subsubsection{Online Hand Candidate Detector}
\textbf{Detector Network Architecture.} In this stage, the SSD \cite{DBLP:journals/corr/LiuAESR15} is chosen as our candidate detector to achieve the balance
between speed and accuracy. The MobileNet \cite{howard2017mobilenets} is used as the backbone CNN network and
the width multiplier (WM) is set to be 0.25 \cite{howard2017mobilenets} to improve the inference speed.
The multi-scale pyramid feature maps are produced by the max pooling operation similar to PPN \cite{DBLP:journals/corr/abs-1807-03284}, for reducing parameters.


\textbf{Training Loss of the Candidate Detector.} 
The overall loss function is a weighted sum of the confidence loss and the regression loss 
\begin{equation}
  \begin{aligned}
    L_{gen} = \alpha_{box}L_{box} + \alpha_{conf}L_{conf} \\
  \end{aligned}
\end{equation}
where $L_{box}$ is the regression loss of bounding boxes following the concept of  \cite{girshick2015fast,DBLP:journals/corr/LiuAESR15},
and the confidence loss $L_{conf}$ is the cross-entropy loss 
with the ground-truth label $y$ and the predicted candidate probability $p$
$$L_{conf}=-\sum_{i}(y_{i}log(p_{i})+(1-y_{i})(1-log(p_{i})))$$

\subsubsection{Hand Detection and Keypoints Regression using the Multi-task CNN}
\textbf{Multi-task Network Architecture.} A multi-task CNN is used as the validation approach to reject erroneous candidates and regress the hand keypoints at the same time. Two branches of fully connected layers (FC) are added to the end of the network for the hand confidence prediction and keypoints regression separately. 

\textbf{Training Loss of the Multi-task Network.} 
The complete loss function of the multi-task network is composed of the cross-entropy 
loss over binary classification and keypoints Euclidean loss as
\begin{equation}
  \begin{cases}
    \begin{aligned}
    L_{kpt} &= \sum_{i}{\left \| \mathbf{u}_{i}^{gt} - \mathbf{u}_{i}^{r} \right \|^{2}} \\
      L_{cls} &= - \sum_{i}{y_{i}log(\bar{p_{i}})} \\
      L_{joint} &= \alpha_{cls}L_{cls} + \alpha_{kpt}L_{kpt}
    \end{aligned}
  \end{cases}
\end{equation}
where $\mathbf{u}=(u,v)^{T}$ is the corresponding coordinate of hand keypoint and $\bar{p}$ is the predicted 
probability of hand.

\subsubsection{Hand Trace Mapping via the Predicted Locations}
In the case of HGR system used in complicated environments, ambiguity caused by multiple hands is normal rather than incidental. While other methods \cite{gupta2016online} often ignore this problem, we here introduce hand trace mapping to solve this problem.
 
\textbf{Hand Trace Mapping.} 
Hand mapping approach is used to match the predicted hand of the current frame with the maintained traces $\mathbb{S}_{trace}$ using the current bounding box and keypoints. We adopt
the match loss $L_{match}$ to measure the similarity of the current hand and existing traces $\mathbb{S}_{trace}$, which combines the intersection over union (IOU) and keypoint locations. 
The overall loss function of the matching phase is defined as 
\begin{equation}
  \begin{aligned}
    L_{match} = \omega_{loc}L_{loc} + \omega_{IoU}L_{IoU} + \omega_{area}L_{area} \\
  \end{aligned}
\end{equation}
where IoU loss $L_{IoU}$ and the region area loss $L_{area}$ of hand bounding box
indicate the regional similarity. The Euclidean loss $L_{loc}$ is utilized to measure the similarity of 
keypoint positions.

\textbf{Temporal Motion Features.} Once a detected hand is associated with the trace $\mathbb{S}_{map}^{k}=[X^{t0}, X^{t1},...,X^{tn}]^T$, the temporal 
motion features $X_{mot}^{t}$ of this tracked hand can be directly computed using the location vector
$X_{p}^{t} = [x_{p0}^{t}, x_{p0}^{t}, ..., x_{pj}^{t}, x_{pj}^{t}]$ as follows:
\begin{equation}
  \begin{cases}
    \begin{aligned}
      X_{v}^{t} &= x_{pi}^{t} - x_{pi}^{t-1} \\
      X_{e}^{t} &= x_{pi}^{t} - x_{pj}^{t}, (pi,pj) \in \mathbb{E} \\
      X_{mot}^{t} &= [X_{v}^{t}, X_{e}^{t}] \\
    \end{aligned}
  \end{cases}
\end{equation}
where $X_{v}^{t}=[x_{v0}^{t}, x_{v0}^{t}, ..., x_{vj}^{t}, x_{vj}^{t}]$ denotes the velocity of the tracked hand.  
$X_{e}^{t}$ indicates the edge vector of hand skeleton, where $\mathbb{E}$ is the edge set. $X_{e}^{t}$ describes the hand shape information.

\subsection{HGR using Temporal Motion Features}

\subsubsection{Online Recognition based the TraceSeqNN}

As depicted in \autoref{fig:TraceSeqNN}, the proposed TraceSeqNN is primarily consisted of transformer, LSTM block and 
FC block. 
Given timestep $T$, the sequential input $X_{mot}^{T}=[X_{mot}^{tn-T}, X_{mot}^{tn-T+1},...,X_{mot}^{tn}]^T$ is fed into a Transformer to reshape the motion features for 
the LSTM layers \cite{hochreiter1997long}. 

In temporal movements, the hand shape and the velocity of keypoints demonstrate different characteristics.
We perform velocity branch and shape branch to learn the velocity features $X_{v}^{T}$ and hand shape features $X_{e}^{T}$ separately.
The outputs of these two branches are further stacked into the FC layers, the probability of gesture categories is finally predicted via the softmax layer.

\subsubsection{Sequence Sample Generation and Augmentation}
\textbf{Sample Generation Principle.}
Given the start and end timestamps $\phi_{s}, \phi_{e}$ of annotated segments in the captured video, 
we can generate the sample set for training and test phases by clipping the original video with an 
objective timestep $T_{obj}$. 
\begin{table}   
  \caption{\textbf{The hand detection results.} 
  The top two rows show the impact of image input size.
  The proposed cascaded network is effective in improving the performance for low-resolution input}
  \label{tab:detection_res}
  \newcommand\ChangeRT[1]{\noalign{\hrule height #1}}
  \newcommand\T{\rule{0pt}{4ex}}
  \newcommand\LH{\rule{0pt}{4ex}}
  \newcommand\B{\rule[-2ex]{0pt}{0pt}}
  \newcommand\BH{\rule[-2ex]{0pt}{0pt}}
  \newcommand{\tabincell}[2]{\begin{tabular}{@{}#1@{}}#2\end{tabular}}
  \renewcommand{\arraystretch}{1}
  \begin{center}
    \resizebox{\linewidth}{!}{
      \begin{tabular}{l|c|cc|cc|c}
     \ChangeRT{1pt}
      \multirow{2}{*}{\textbf{\LARGE Model}} &  \multirow{2}{*}{\textbf{\LARGE Backbone}}
      & \multicolumn{2}{c|}{\textbf{\LARGE Our Dataset}}  & \multicolumn{2}{c|}{\textbf{\LARGE Nvidia}} 
      &  \multirow{2}{*}{\textbf{\tabincell{c}{\LARGE FPS \\ \large (MTK8167 CPU)}}} \T\LH \\ 
      \cline{3-4} \cline{5-6} 
      & & \LARGE Recall & \LARGE Precision & {\LARGE Recall} & {\LARGE Precision} & \T\B \\ 
      \hline 
      \tabincell{lc}{\LARGE SSD} &  \tabincell{c}{\LARGE MobileNet-V1 \\ \large (300x300,WM:1.0)} & \LARGE $98.77\%$ & \LARGE $94.64\%$ & \LARGE $96.55\%$ & \LARGE $95.82\%$ & \LARGE $4.5$  \LH\BH \\ 
      \tabincell{lc}{\LARGE SSD} &  \tabincell{c}{\LARGE MobileNet-V1 \\ \large (224x224,WM:0.25)} & \LARGE $91.54\%$ & \LARGE $90.85\%$ & \LARGE $93.12\%$ & \LARGE $94.84\%$ & \LARGE $19$ \LH\BH  \\
      \tabincell{lc}{\LARGE Propsed} & \tabincell{c}{\LARGE MobileNet-V1 \\  \large (224x224,WM:0.25)} & \LARGE $94.46\%$ & \LARGE $93.10\%$ & \LARGE $94.78\%$ & \LARGE $98.89\%$&  \LARGE $16$ \LH  \\ 
      \ChangeRT{1pt}
      \end{tabular}}
    \end{center}
  \end{table}

\begin{table}   
  \caption{\textbf{Evaluation the regression errors on public dataset.} 
  We conduct experiments on public datasets to compute the error metrics for keypoint regression.
}
  \label{tab:keypoints_metric}
  \newcommand\ChangeRT[1]{\noalign{\hrule height #1}}
  \newcommand{\tabincell}[2]{\begin{tabular}{@{}#1@{}}#2\end{tabular}}
  \newcommand\T{\rule{0pt}{4ex}}
  \newcommand\LH{\rule{0pt}{3ex}}
  \newcommand\B{\rule[-2ex]{0pt}{0pt}}
  \newcommand\BH{\rule[-2ex]{0pt}{0pt}}
  \renewcommand{\arraystretch}{1}
  \begin{center}
    \resizebox{\linewidth}{!}{
      \begin{tabular}{l|c|c|cc|c}
      \ChangeRT{1pt}
      \multirow{2}{*}{\textbf{ \LARGE Models}} & \multirow{2}{*}{\textbf{ \LARGE Dataset}} & \multirow{2}{*}{\textbf{ \LARGE \tabincell{c}{Dataset \\ Resolution}}}
      & \multicolumn{2}{c|}{\textbf{ \LARGE Error}} & \multirow{2}{*}{\textbf{ \tabincell{c}{\LARGE FPS \\ \large MacProI7-CPU}}} \T\B\\
      \cline{4-5} 
      & & & {\LARGE Mean} & {\LARGE Median} \LH\BH \\ 
      \hline  
      
       \multirow{2}{*}{\textbf{ \tabincell{lc}{\LARGE Ours \\  \large keypoints Regression}}} & \tabincell{lc}{\LARGE Nvidia \cite{gupta2016online}} &  \LARGE $320x240$  & \LARGE $12.74$ &  \LARGE $9.85$  & \LARGE 278 \LH\BH\\
       & \tabincell{lc}{\LARGE Our dataset } &  \LARGE $1280x720$ & \LARGE $9.47$ &  \LARGE $7.61$ &\LARGE  278 \LH\BH\\
       \hline
       \multirow{2}{*}{\textbf{  \tabincell{lc}{\LARGE CPM-1Stage}}} & \tabincell{lc}{\LARGE Nvidia \cite{gupta2016online}} &  \LARGE $320x240$  & \LARGE $10.52$ & \LARGE  $5.38$ & \LARGE 91 \LH\BH \\
       & \tabincell{lc}{ \LARGE Our dataset } &\LARGE  $1280x720$ & \LARGE  $7.06$ & \LARGE $3.16$ &\LARGE  91 \LH\BH \\
       
      \ChangeRT{1pt}
      \end{tabular}}
    \end{center}
  \end{table}

\begin{table*}   
  \caption{\textbf{Evaluation the TraceSeqNN for HGR on our dataset.} Comparing our models against different state-of-the-art approaches 
  on the same sequence dataset, which consists of left waving, right waving, and negative samples. 
  Taking our temporal motion features as input, the HGR can run at 250 fps on low-cost MediaTek MTK-8167S processor.
  }
  \label{tab:traceseq_res}
  \newcommand\ChangeRT[1]{\noalign{\hrule height #1}}
  \newcommand{\tabincell}[2]{\begin{tabular}{@{}#1@{}}#2\end{tabular}}
  \renewcommand{\arraystretch}{1.3}
  \begin{center}
    \resizebox{0.9\linewidth}{!}{
      \begin{tabular}{l|c|c|ccc|ccc|c|c|c}
      \ChangeRT{1pt}
      \multirow{2}{*}{\textbf{Model}} & \multirow{2}{*}{\tabincell{c}{\textbf{DataAug} \\ (temporal)}} 
      & \multirow{2}{*}{\textbf{Input}} & \multicolumn{3}{c|}{\textbf{Recall}} & \multicolumn{3}{c|}{\textbf{Precision}} 
      & \multirow{2}{*}{\textbf{Accuracy}}
      & \multirow{2}{*}{\textbf{False Positive}} & \multirow{2}{*}{{\textbf{\tabincell{c}{Speed \\ \small{(MTK8167 CPU)}}}}} \\
      \cline{4-6} \cline{7-9}
      & & & {neg} & {lwave} & {rwave} & {neg} & {lwave} & {rwave} & & & \\ 
      \hline  
      \tabincell{lc}{TDNN \cite{waibel1995phoneme})} & - & {$X_{mot}^{T}$} & $98.84\%$ & $72.49\%$ & $83.68\%$ & $84.26\%$ & $94.32\%$ & $95.46\%$ & $88.55\%$ & $5.1\%$ & $1ms$ \\ 
      \tabincell{lc}{TDNN \cite{waibel1995phoneme}} & \checkmark & {$X_{box}^{T}$} & $91.09\%$ & $91.91\%$ & $90.45\%$ & $94.61\%$ & $88.30\%$ & $87.32\%$ & $91.15\%$ & $12.2\%$ & $1ms$ \\
      \tabincell{lc}{TDNN \cite{waibel1995phoneme}} & \checkmark & {$X_{mot}^{T}$} & $97.73\%$ & $87.36\%$ & $86.37\%$ & $91.53\%$ & $95.05\%$ & $91.75\%$ & $92.40\%$ & $6.6\%$ & $1ms$ \\
      \tabincell{lc}{LSTM \cite{hochreiter1997long}} & \checkmark & {$X_{mot}^{T}$} & $99.26\%$ & $86.43\%$ & $88.26\%$ & $92.16\%$ & $95.78\%$ & $93.96\%$ & $93.39\%$ & $5.1\%$ & $4ms$ \\ 
      \tabincell{lc}{CLDNN \cite{sainath2015convolutional}} & \checkmark & {$X_{mot}^{T}$} & $97.08\%$ & $88.75\%$ & $90.95\%$ & $94.62\%$ & $92.63\%$ & $91.95\%$ & $93.51\%$ & $7.7\%$ & $4ms$ \\
      \tabincell{lc}{LRCN \cite{donahue2015long}} & \checkmark & {$X_{mot}^{T}$} & $97.77\%$ & $89.41\%$ & $88.75\%$ & $93.77\%$ & $93.67\%$ & $92.72\%$ & $93.50\%$ & $6.8\%$ & $4ms$ \\ 
      \tabincell{lc}{Proposed} & \checkmark & {$X_{mot}^{T}$} & $99.58\%$ & $90.15\%$ & $89.15\%$ & $93.92\%$ & $95.85\%$ & $95.42\%$ & $94.71\%$ & $4.4\%$ & $4ms$ \\ 
      \ChangeRT{1pt}
      \end{tabular}}
    \end{center}
  \end{table*}
In order to distinguish positive and negative samples, we compute the labels of these samples according to the similarity of these clipped samples and their corresponding original segments. Firstly, we represent the similarity as the intersection over a sample $r_{IoS}$ and the intersection over its annotation $r_{IoA}$ 
\begin{equation}
  \begin{cases}
    \begin{aligned}
      I_{overlap} &= \min(\phi_{e},\psi_{e}) - \max(\phi_{s}, \psi_{s}) \\
      r_{IoA} &= I_{overlap}/(\phi_{e}-\phi_{s}) \\
      r_{IoS} &= I_{overlap}/(\psi_{e}-\psi_{s})
    \end{aligned}
  \end{cases}
\end{equation}
where $\psi_{s},\psi_{e}$ are the start and end timestamps of the clipped sequence in the original video.

Subsequently, the label of this clipped sample can be computed according to the similarity metric
\begin{equation}
  label = 
  \begin{cases}
      0, & \text{if } r_{IoS} < \delta_{IoS} \text{ or } r_{IoA} < \delta_{IoA}\\
      c, & \text{otherwise }\\
  \end{cases}
\end{equation}
where $c$ is the annotated category in the original video.
$\delta_{IoS}$ and $\delta_{IoA}$ are the thresholds of the sample principle.

\textbf{Data Augmentation for the Temporal Domain.}
In order to generate massive hand motion samples of different temporal
domains using the limited-quantity annotated segments, we propose a novel data augmentation method
to increase the diversity of temporal domains. Firstly, the sequence samples are yielded via the timestep set:

\begin{equation}
  T_{aug} = \left\{ T|T_{min}+n*\Delta{T}, n\in\mathbb{N}\right\} \\
\end{equation}
where $T_{min}$ is the minimum time step and $\Delta{T}$ is configured to a constant value.

Subsequently, using the nonlinear transformation, i.e. interpolation or down-sampling 
operations, the generated samples are further re-sampled and deformed to the target timestep $T_{obj}$ .


\section{Experiments}
\subsection{Dataset}

\begin{table}   
  \caption{\textbf{Comparison with the state-of-the-art methods on the pubilshed Nvidia dataset.} 
   Proposed-reg uses the multi-task CNN to regress the keypoints, while proposed-cpm adopts the CPM-1Stage network to predict the keypoint locations.
   TraceSeqNN-SB uses the single branch to learn the motion features. 
}
  \label{tab:public_dataset}
  \newcommand\T{\rule{0pt}{2ex}}
  \newcommand\LH{\rule{0pt}{3ex}}
  \newcommand\B{\rule[-2ex]{0pt}{0pt}}
  \newcommand\BH{\rule[-2ex]{0pt}{0pt}}
  \newcommand{\tabincell}[2]{\begin{tabular}{@{}#1@{}}#2\end{tabular}}
  \renewcommand{\arraystretch}{1}
  \begin{center}
    \resizebox{\columnwidth}{!}{
      \begin{tabular}{lccc}
      \toprule[1pt]
      {\textbf{Method}} & {\textbf{Modality}}
     & {\textbf{Accuracy}} & {\textbf{\tabincell{c}{FPS \\ \small{(MacProI7-CPU)}}}} \T\B \\ 
      \hline 
      \tabincell{lc}{Spatial stream CNN \cite{simonyan2015two}} & color & $54.6\%$ & $1$  \LH\BH \\ 
      \tabincell{lc}{C3D \cite{molchanov2015hand}} & color & $69.3\%$ & $1.5$  \LH\BH \\ 
      \tabincell{lc}{R3DCNN \cite{gupta2016online}} & color & $74.1\%$ & $0.15$ \LH\BH  \\
      \hline
      \tabincell{lc}{Proposed-reg  \\ \small (TraceSeqNN-SB)} &color & $65.2\%$ & $76$ \LH\BH  \\
      \tabincell{lc}{Proposed-reg, Data-Aug \\ \small (TraceSeqNN-SB)} &color & $70.1\%$ & $76$ \LH\BH  \\
      \tabincell{lc}{Proposed-reg, Data-Aug \\ \small (TraceSeqNN)} &color & $75.3\%$ & $76$ \LH\BH \\ 
      \tabincell{lc}{Proposed-cpm, Data-Aug \\ \small (TraceSeqNN)} &color & $77.2\%$ & $34$ \LH\BH \\ 
      \bottomrule[1pt]   
      \end{tabular}}
    \end{center}
  \end{table}
  
\subsubsection{Public Nvidia Gesture Dataset}
Nvidia corporation published a dataset of 25 gesture types intended for touchless interfaces in cars\cite{gupta2016online}.
The dataset consists of a total of 1050 training and 482 testing video sequences, covering both bright and dim artificial lighting.
In order to validate the effectiveness of our proposed algorithm on Nvidia dataset, we also add annotations for the keypoints and
bounding boxes of this dataset during the training phase. All new annotations
will be publicly available. 

\subsubsection{Our Dataset}
In addition to testing in the public Nvidia dataset, we also build our own dataset in a crowd-sourcing way.
This dataset consists of more than 150k RGB images 
in different environments conditions (e.g., different background, lighting conditions, and so on) via different RGB cameras, which contains 120k training images and 30k testing. 
In addition, we recorded 30000 video sequences using different RGB cameras at 30fps.

\subsection{Evaluation of Multi-task Cascaded Network for Hand Detection}
The proposed cascaded multi-task network is adopted to predict the hand confidence and regress the keypoint locations. 
In this section,
we conduct experiments for quantitatively evaluating the 
metrics of our architecture design, 
based on an hand-held low-cost interactive device. The main processor on this device is 
MediaTek MTK-8167S. 

\subsubsection{Training Details}
To train the proposed network,
we annotated 8867 ground-truth bounding boxes for Nvidia dataset and 90000 ground-truth bounding boxes for our dataset. The candidate detector was trained with the annotated datasets.
Subsequently, for the multi-task CNN, we generated the positive samples with IOU over 0.5 and negative samples with IOU less than 0.3 via the candidate detector model.
The hand keypoints of the positive samples are annotated only for the training phase of multi-task CNN. 

\subsubsection{Evaluation the Multi-task Cascaded Network}
We compared the proposed network against standard SSD to evaluate the accuracy on previously mentioned datasets. 
For fair comparison, we chosen the MobileNetV1 as the backbone of our detector and standard SSD.
Results from \autoref{tab:detection_res} show that the detection results are drastically degraded with SSD when using low-resolution input. 
However, the proposed network greatly improves both detection recall and precision
with only negligible computational cost.

We also measured the error metrics on our dataset and 
Nvidia \cite{gupta2016online} dataset to calculate the precision of regression. The results are shown in \autoref{tab:keypoints_metric}. Furthermore, we implemented a one-stage convolutional pose machine (CPM-1Stage) \cite{wei2016convolutional} using the MobileNet as the backbone. 

\subsection{Evaluation the TraceSeqNN and Data Augmentation}
\subsubsection{Training Details}
For data augmentation of the training set, we set the objective timestep $T_{obj}$ to be 13 according to the distribution of $\psi_{s}$ and $\psi_{e}$. The minimum timestep $T_{min}$ is set to be 8 and $\Delta{T}$ is 5.
Then, $\delta_{IoA}$ and $\delta_{IoS}$ are set to be 0.3 and $\delta_{TL}$ is 0.85 for calculating labels of the clipped samples.
During the training phase of TraceSeqNN, the initial learning rate is set to be 0.004.
The number of LSTM hidden layers is set to be 64.
We train all the models using Adam Optimizer \cite{kingma2014adam} with the dropout rate
of LSTM cells at 0.2.

\subsubsection{Metrics on Our Dataset using TraceSeqNN}

We designed comprehensive experiments to evaluate different input channels, different data augmentation methods, and different network architectures (i.e., TDNN \cite{waibel1995phoneme} 
, LSTM \cite{hochreiter1997long}, CLDNN \cite{sainath2015convolutional} and LRCN \cite{donahue2015long}) of our TraceSeqNN. 
The results are shown in \autoref{tab:traceseq_res}.
The results demonstrate that our data augmentation method can significantly improve the overall prediction accuracy.
Compared with $X_{box}^{T}$, by using motion features $X_{mot}^{T}$ as the network input, false positives can be largely eliminated. 
To summarize, the proposed TraceSeqNN achieves the highest accuracy for HGR.

\begin{figure}
    \centering
    \includegraphics[scale=0.32]{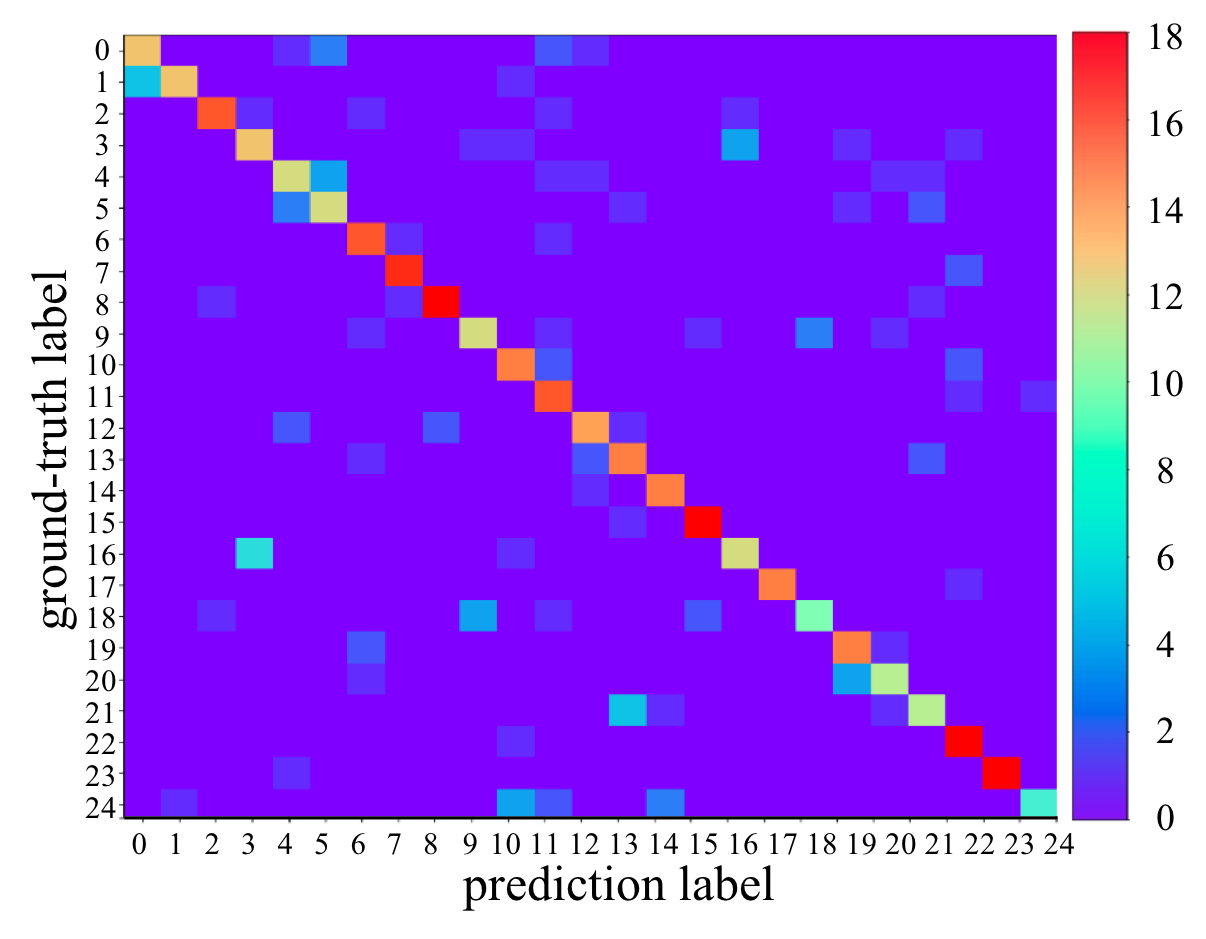}
    \caption{\textbf{Confusion matrix of 25 gesture types.} Confusion matrix on the total 482 Nvidia testset.}
    \label{fig:confusion_matrix}
    \end{figure}
    
\begin{figure}
\centering
\includegraphics[scale=0.33]{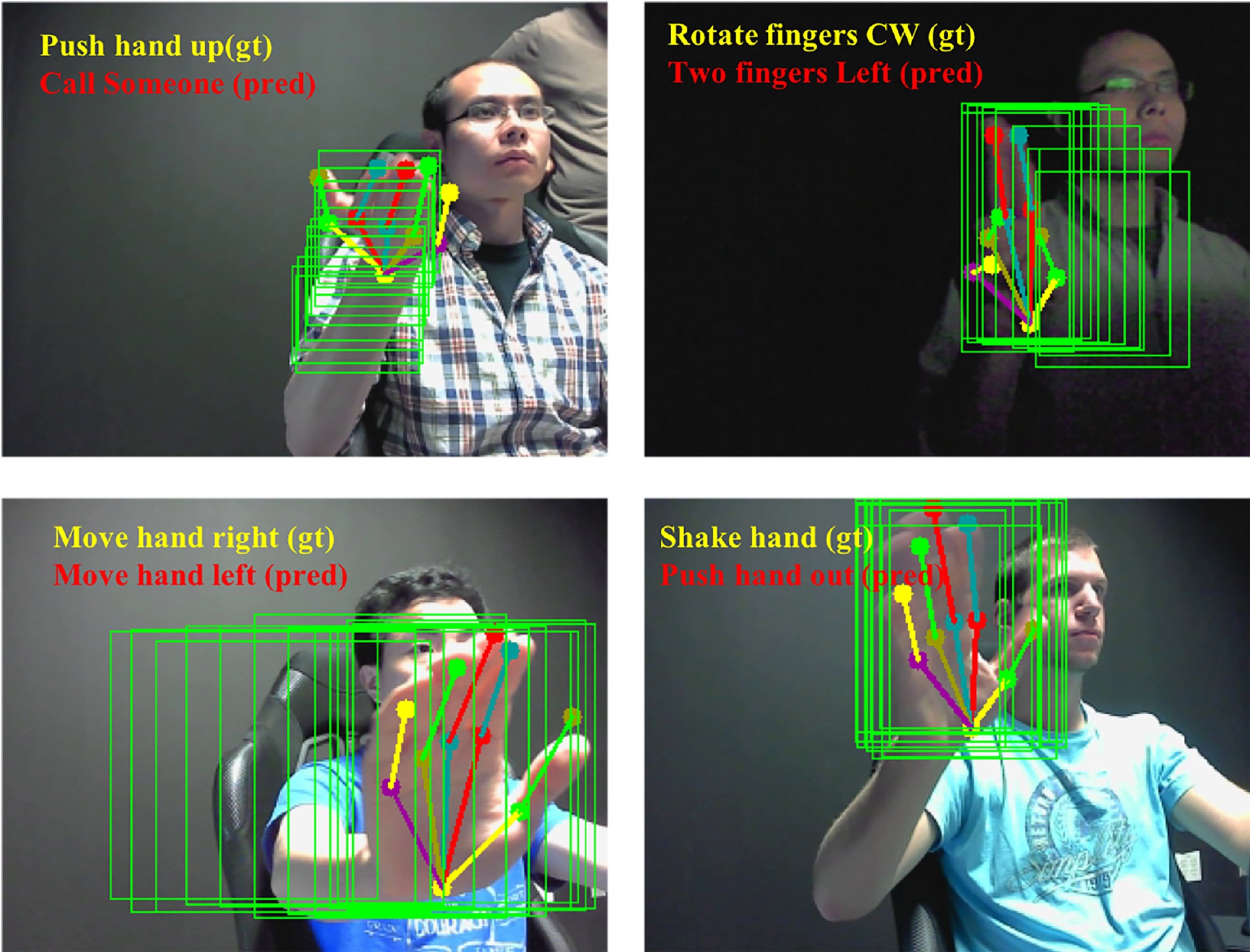}
\caption{\textbf{Examples of failure cases.} We show failure cases due to ambiguity of hand gesture or overlapping between gestures, e.g. "Two fingers left" is part of "Rotate fingers CW" gesture.}
\label{fig:aliha ondset}
\end{figure}

\subsection{Evaluation on the Public Datasets for HGR}

\autoref{tab:public_dataset} shows the results by comparing our approach against competing state-of-the-art networks on the Nvidia dataset.
As some of the ops is not supported on the hand-held low-cost device, for ensuring fair comparison, all the experiments were performed on a MacBook Pro core-i7 computer by only using CPU 
and 16GB memory. 
\autoref{tab:public_dataset} demonstrates that C3D \cite{molchanov2015hand} and R3DCNN \cite{gupta2016online} rely heavily on computational resources, which can only run at $1$fps.

For the proposed TraceSeqNN, 
benefiting from the designed lightweight cascaded framework, our LE-HGR significantly improves the 
inference speed at $76$ fps. In addition, the proposed approach achieves the highest accuracy over competing methods using temporal motion
features. Adopting our proposed data augmentation, we have seen an increase of $5\%$ on this challenging gesture dataset.
Our final accuracy is $75.3\%$ by only using the RGB modality dataset.

The complete confusion matrix is shown in \autoref{fig:confusion_matrix}.
If equipped with a more precise keypoints predictor (CPM-1Stage), we can get a $2\%$ improvement. we show some failure cases in \autoref{fig:aliha ondset}

\subsection{Typical AR applications}

In general, AR devices have limited user interfaces, most often small buttons or touchscreen, these interfaces are not natural for human and destroy AR immersion experiences.

\begin{figure}
    \centering
    \includegraphics[scale=0.3]{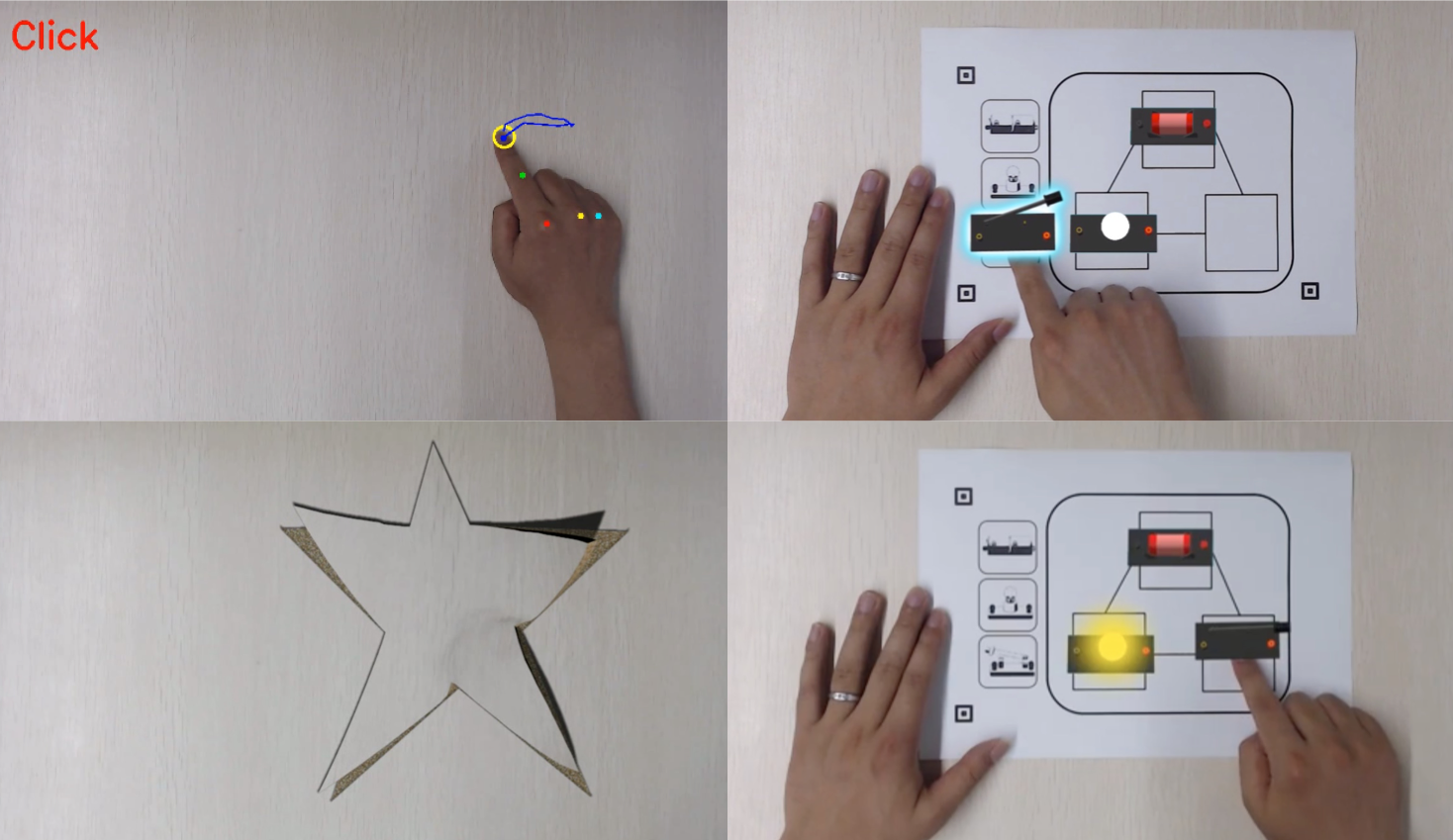}
    \caption{\textbf{Typical AR application.} The left column shows that finger track and gesture recognition ("click") via our LE-HGR approach. The right column shows the application of AR
    education.}
    \label{fig:gesture_ar}
    \end{figure}

In order to enrich the AR experiences, we design an online AR interacting system via direct hand touch and gesture sensing. 
As shown in the \autoref{fig:gesture_ar}, by locating and tracking finger position, it can understand the user's intention (e.g. "clicking", "draw star", and so on). Obviously, by applying this system to educational and entertainment applications, we can provide more incredible and interesting AR experiences. We present a video to show these AR application demos in our supplementary material.



\section{Conclusion}
We have presented a complete online RGB-based gesture recognition framework, 
which is extremely lightweight and efficient for low-cost commercial embedded devices.
An improved multi-task cascaded network is introduced for online hand detection and a prototype of the hand trace mapping is implemented to tackle the interference of multi-hand.
Besides, a TraceSeqNN network combined with an effective data augmentation method is performed to learn the temporal information and recognize the gesture categories.

The experimental results show that our proposed cascaded multi-task network significantly improves the recall and 
precision rates of hand detection. The designed LE-HGR framework achieves advanced accuracy with significantly reduced computational complexity. 
To extend this work, we are interested in investigating this challenging issue directly to predict the 3D hand keypoint using only RGB-cameras.
Moreover, a promising future research direction is to explore the 
attention mechanism for RGB-based HGR applications.


\bibliographystyle{abbrv-doi}

\bibliography{handgesture}

\end{document}